\documentclass[letterpaper]{article} 
\usepackage{aaai25}  
\usepackage{times}  
\usepackage{helvet}  
\usepackage{courier}  
\usepackage[hyphens]{url}  
\usepackage{graphicx} 
\usepackage{natbib}  
\usepackage{caption} 
\usepackage{algorithm}
\usepackage{algorithmic}

\usepackage{newfloat}
\usepackage{listings}
\DeclareCaptionStyle{ruled}{labelfont=normalfont,labelsep=colon,strut=off} 
\floatstyle{ruled}
\newfloat{listing}{tb}{lst}{}
\floatname{listing}{Listing}
\usepackage{amsthm}
\newtheorem{definition}{Definition}
\usepackage{amsmath}
\usepackage{amssymb}
\usepackage{amsfonts}
\usepackage{subfig}
\usepackage{multirow}
\usepackage{booktabs}
\usepackage{diagbox}
\usepackage{xspace}
\newcommand{\ours}{\textsc{NBC}\xspace}
\newcommand{\ourloss}{l^{\textsc{NBC}}}
\usepackage{mdframed}
\usepackage{makecell}
\newmdenv[
  linecolor=black,
  linewidth=1pt,
  roundcorner=5pt,
  backgroundcolor=white,
  innerleftmargin=5pt,
  innerrightmargin=5pt,
  innertopmargin=5pt,
  innerbottommargin=5pt,
]{custombox}

\title{Training Verification-Friendly Neural Networks via Neuron Behavior Consistency}
\author{
    Zongxin Liu\textsuperscript{\rm 1,\rm2},
    Zhe Zhao\textsuperscript{\rm 3},
    Fu Song\textsuperscript{\rm 1,\rm 4},
    Jun Sun\textsuperscript{\rm 5},
    Pengfei Yang\textsuperscript{\rm 6},
    Xiaowei Huang\textsuperscript{\rm 7},
    Lijun Zhang\textsuperscript{\rm 1,\rm2}\thanks{Corresponding author.}
}
\affiliations{
    \textsuperscript{\rm 1}Key Laboratory of System Software (Chinese Academy of Sciences) and State Key Laboratory of Computer Science, Institute of Software, Chinese Academy of Sciences, Beijing, China\\
    \textsuperscript{\rm 2}University of Chinese Academy of Sciences, Beijing, China\\
    \textsuperscript{\rm 3}RealAI, Beijing, China\\
    \textsuperscript{\rm 4}Nanjing Institute of Software Technology, Nanjing, China\\
    \textsuperscript{\rm 5}Singapore Management University, Singapore\\
    \textsuperscript{\rm 6}College of Computer and Information Science, Software College, Southwest University, Chongqing, China\\
    \textsuperscript{\rm 7}The University of Liverpool, Liverpool, United Kingdom\\
    liuzx@ios.ac.cn,
    zhe.zhao@realai.ai,
    songfu@ios.ac.cn,
    junsun@smu.edu.sg,
    ypfbest001@swu.edu.cn,
    xiaowei.huang@liverpool.ac.uk,
    zhanglj@ios.ac.cn
}

\usepackage{bibentry}

\begin{document}

\maketitle

\begin{abstract}
Formal verification provides critical security assurances for neural networks, yet its practical application suffers from the long verification time. This work introduces a novel method for training verification-friendly neural networks, which are robust, easy to verify, and relatively accurate. Our method integrates neuron behavior consistency into the training process, making neuron activation states remain consistent across different inputs within a local neighborhood. This reduces the number of unstable neurons and tightens the bounds of neurons thereby enhancing the network's verifiability. We evaluated our method using the MNIST, Fashion-MNIST, and CIFAR-10 datasets with various network architectures. The experimental results demonstrate that networks trained using our method are verification-friendly across different radii and architectures, whereas other tools fail to maintain verifiability as the radius increases. Additionally, we show that our method can be combined with existing approaches to further improve the verifiability of networks.

\end{abstract}

\section{Introduction}

Neural networks are increasingly being applied in safety-critical domains such as autonomous driving~\cite{selfdriving} and flight control~\cite{julian2019deep}. However, they often struggle with a lack of robustness, as even minor perturbations to their inputs can lead to incorrect predictions~\cite{BuZDS22,ChenCFDZSL21,SongLCFL21,ChenZZS23,CZSCFL22,ZCWYSS21,chen2022towards}. This is unacceptable in safety-critical applications, where consistent performance is imperative. 
Thus, it is desirable to develop methods to systematically advance the robustness verification of neural networks.

Existing methods for analyzing robustness can be broadly classified into two categories: empirical analysis through adversarial attacks and mathematical proof via formal verification. While adversarial attacks generate misleading examples, they merely demonstrate the presence of adversarial samples without affirming their absence. In contrast, formal verification ensures the correctness of neural networks using logical and mathematical methods, allowing us to verify their robustness formally. This rigorous verification is essential for safety-critical systems.

Advanced formal verification tools~\cite{wang2021beta,zhang2022general,bak2021nnenum,reluplex}, typically employ branch-and-bound algorithms for neural network verification.
At the start of the verification process, abstract interpretation~\cite{gehr2018ai2,mirman2018differentiable,ZhangZCSC21,ZhangZCSC23,GuoWZZSW21} is usually used to abstract neurons.
If the properties of the network remain undetermined after using symbolic propagation~\cite{deeppoly} to calculate neuron boundaries and applying MILP~\cite{tjeng2017evaluating,tran2020nnv,ZhangZCSZCS22,ZhangSS23,ZhangCSSD24} or SMT methods~\cite{ehlers2017formal,huang2017safety,reluplex,ZhaoZCSCL22,LiuXSSXM24} for constraint solving, further branching is required. 
The branching process involves enumerating the activation states of unstable neurons, whose activation status cannot be determined through bound calculations. This introduces additional constraints, thereby refining the abstraction.
Typically, neural networks contain numerous unstable neurons, and exploring the combinations of these activation states requires exponential time, which limits the widespread application of formal verification techniques in practice.

In addition to developing ever-more sophisticated methods for post-training verification, researchers have investigated the idea of training neural networks that are easier to verify, known as verification-friendly neural networks.
Ideally, a training method for verification-friendly neural networks must satisfy the following requirements. 
First, (accuracy) the resultant neural network must have an accuracy comparable to that of neural networks trained conventionally. 
Second, (robustness) the resultant neural network must have improved robustness, which could be measured using existing adversarial attacks. 
Third, (verifiability) 
it must be easier to verify using existing or dedicated neural network verification techniques, which can be measured using the effectiveness of
selected neural network verification methods.

There are mainly two existing approaches to promoting the verification-friendliness of neural networks. One approach involves post-processing the network using methods that modify the weights of the networks~\cite{relustable2019,baninajjar2023verification}. 
The other involves altering the neural network design and training process with considerations for verification~\cite{relustable2019,narodytska2019search,training_for_verification}.
However, these methods still have limitations. 
The effectiveness of post-training methods is limited by the network itself. Additionally, certain post-training methods such as using MILP to make the network sparser~\cite{baninajjar2023verification}, are computationally expensive and may not scale well for large networks. 
The ReLU Stable methods, which introduce ReLU Stable (RS) loss~\cite{relustable2019} to reduce the number of unstable neurons, 
depend on the bounds of neurons. When the perturbation radius changes, it often causes these bounds to shift, affecting the verification efficiency of the network.
Certified training, which is primarily based on the heavy Interval Bound Propagation (IBP) method~\cite{mirman2018differentiable,gowal2018effectiveness,zhang2019towards,xu2020automatic} suffers from long training times and issues like gradient explosion or vanish problem. 
Adversarial training~\cite{DBLP:MadryMSTV18, trades,ganin2016domain,zhu2917upaired} typically increases network robustness but does not contribute to improving its verifiability.

In this work, we introduce a straightforward yet effective training method that enhances the verifiability of neural networks by enforcing the consistency of neuron behavior, which we refer to as neuron behavior consistency (NBC), throughout the training process as a regularization term. A neuron is called behavior consistent if its activation state remains the same within a given input neighborhood. By maximizing the consistency of neurons, the unstable neurons are decreased, reducing the search space of the verification process. NBC can also help tighten the bounds of neurons, as fewer unstable neurons introduce less error during bound calculation algorithms.
Our approach can be scaled to larger networks compared to MILP-based methods. Moreover, the core of our method lies in the consistency of neuron behavior without relying on heavy IBP methods, which reduces training epochs and ensures that the trained network maintains verifiability across different perturbation radii.

We evaluate our method using Fashion-MNIST, MNIST, and CIFAR-10 datasets across different architectures at various perturbation radii. Our method outperforms others in stable neuron ratio and achieves up to a 450\% speedup in verification time.
Importantly, our method accelerates the verification while preserving the accuracy of the models, which is not commonly achieved by existing methods. 
In summary, our contributions are as follows:
\begin{itemize}
    \item We introduce a method of training verification-friendly networks by integrating neuron behavior consistency.
    \item We evaluate our method on three well-known datasets.
    Experimental results show that networks trained using our method can maintain verification-friendly properties across different radii and different model architectures.
    \item We demonstrate that our method can be combined with existing methods to further improve the verifiability of networks, especially in the case of large networks.
    \item We show that our method accelerates the verification process while preserving model accuracy and robustness.
\end{itemize}

\section{Preliminary}\label{sec: preliminary}

In this section, we introduce the background of neural network verification problems and the general branch and bound verification framework.

\subsection{Neural Networks Verification Problems}

Given a neural network $f: \mathbb{R}^{m_{\mathrm{in}}} \rightarrow \mathbb{R}^{m_{\mathrm{out}}}$, with $m_{\mathrm{in}}$ input neurons and $m_{\mathrm{out}}$ output neurons, the goal of the neural network verification problem is to determine whether the output of the network satisfies a set of output constraints $\mathcal{P}$ for all inputs that meet the input constraints $\mathcal{C}$, formally defined as:
\begin{definition}[Neural Network Verification Problem]
The neural network verification problem $\langle f, \mathcal{C}, \mathcal{P} \rangle$ is to determine whether: 
\begin{equation}
    \forall \boldsymbol{x} \in \mathbb{R}^{m_{\mathrm{in}}}, \boldsymbol{x} \in \mathcal{C} \Rightarrow f(\boldsymbol{x}) \in \mathcal{P},
\end{equation}
where $\mathcal{C}\subseteq \mathbb{R}^{m_{\mathrm{in}}}$ represents the input constraints and $\mathcal{P}\subseteq \mathbb{R}^{m_{\mathrm{out}}}$ represents the output constraints.
\end{definition}

We focus on the local robustness verification problem $\langle f, \mathcal{C}_\epsilon(\boldsymbol{x}), \mathcal{P}_{c} \rangle$, which checks if the classification result $c$ is robust to input perturbations $\varepsilon$ in the $l_{\infty}$-norm, where the input constraints $\mathcal{C}_\epsilon(\boldsymbol{x})$ are defined as $\{\boldsymbol{x} \in \mathbb{R}^{m_{\mathrm{in}}} \mid |\boldsymbol{x} - \boldsymbol{x}^0|_{\infty} \leq \varepsilon \}$ and the output constraints $\mathcal{P}_{c}$ are defined as $\{\boldsymbol{y} \in \mathbb{R}^{m_{\mathrm{out}}} \mid \bigwedge_{i \neq c} \boldsymbol{y}_i - \boldsymbol{y}_{c} \leq 0 \}$.

\subsection{General Verification Framework}\label{sec:verification_framework}

State-of-the-art methods for solving neural network verification problems~\cite{wang2021beta,zhang2022general, marabou,bak2021nnenum} are typically based on branch-and-bound algorithms, consisting of three critical components: constraint solving, bound calculation, and branch selection.

\begin{figure}[t]
    \centering
    \includegraphics[page=1]{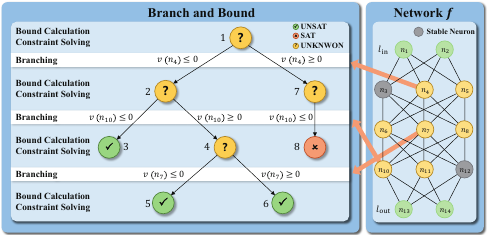}
    \caption{The Branch and Bound (BaB) verification process}
    \label{fig: verification_framework}
\end{figure}

As shown in Figure~\ref{fig: verification_framework}, the verification process begins with the bound calculation, where the upper and lower bounds of neuron outputs are estimated under specified input constraints. If these bounds are sufficiently precise, the properties of the network can be directly verified. Due to non-linear activation functions such as ReLU, computing these bounds can be complex, which is a problem often addressed through neuron-wise abstraction~\cite{deeppoly,bak2021nnenum}. This abstraction uses linear bounds to approximate neuron outputs, thereby simplifying the bound calculations. 

When the bound calculation does not suffice to verify the network's properties, constraint solving is employed. This process involves determining if the constraints of the abstracted network can be satisfied, using Linear Programming (LP) or Mixed Integer Linear Programming (MILP) methods. Constraints typically include the input constraints $\mathcal{C}$, the negated output constraints $\neg\mathcal{P}$, and the constraints of the abstracted network itself.
If these constraints are UNSAT (unsatisfiable), the property holds, proving that 
the network can not be successfully attacked within $\mathcal{C}$;
otherwise, the returned counterexample must be examined. If the counterexample is not a false positive caused by over-approximation, it indicates a violation of the network's properties. Otherwise, it suggests that the network is over-abstracted, necessitating branch selection to refine the abstraction.

Branch selection selects an unstable neuron (whose activation state cannot be determined through bound calculations) and splits it into active and inactive branches. Identifying the state of a neuron entails establishing new constraints that refine the abstraction. 
Branch selection strategies significantly affect verification efficiency. While branching may require the exploration of all unstable neurons in worst-case scenarios, minimizing these neurons can exponentially reduce the theoretical upper bound on branching.

\section{Training Verification-Friendly Networks via Neuron Behavior Consistency}\label{sec: verification_friendly_neural_networks}

In this section, we propose our method, neuron behavior consistency, to train verification-friendly neural networks.  

Given a network $f$ with parameters $\theta$ and an underlying distribution $\mathcal{D}$, the traditional training process aims to optimize the parameters $\theta$ to minimize the expected loss:
{\fontsize{8pt}{8pt}\selectfont
\begin{equation} 
    \min_{\theta} \underset{(\boldsymbol{x}, {y}) \sim \mathcal{D}}{\mathbb{E}} \left({loss}(\boldsymbol{x}, \boldsymbol{y})\right),
\end{equation}
}where $(\boldsymbol{x}, {y})$ denotes the input and target label sampled from the distribution $\mathcal{D}$, and $\boldsymbol{y}$ is the one-hot vector representation of the target label $y$, that is, a vector with a single 1 at the index of the target label and 0 elsewhere. The most common loss function is the cross-entropy function for classification tasks, defined as $\text{CE}(f(\boldsymbol{x}), \boldsymbol{y}) = -\sum_{i} \boldsymbol{y}_i \log f(\boldsymbol{x})_i$.

The ordinary training objective does not impose constraints on neurons, potentially resulting in a large number of unstable neurons. 
We thus propose an alternative training objective that aims to maximize the consistency of neurons. A neuron is called consistent if its activation state remains the same within a given input neighborhood. Formally, given an input $\boldsymbol{x}$ and a neighboring input $\boldsymbol{x}'$, the consistency of the $j$-th neuron in the $i$-th layer $n^{(i)}_j$ is defined as:

{\fontsize{8pt}{8pt}\selectfont
\begin{equation} \label{eq: original_nbc}
    {\ours}(n^{(i)}_j, \boldsymbol{x}, \boldsymbol{x}') = \begin{cases}
        1, & \text{if } \mathrm{sign}(f^{(i)}(\boldsymbol{x})_j) = \mathrm{sign}(f^{(i)}(\boldsymbol{x}')_j), \\
        0, & \text{otherwise},
    \end{cases}
\end{equation}
}

\noindent
where $f^{(i)}(\boldsymbol{x})_j$ denotes the pre-activation value of the $j$-th neuron of the $i$-th layer when fed with input $\boldsymbol{x}$.
Intuitively, for any input within a given neighborhood, if the activation states of individual neurons are highly consistent, the calculated boundaries (upper and lower bounds) are more likely to be tight. This coherence may reduce the occurrence of unstable neurons in the neural network.

To maximize (minimize the negative) this consistency, we incorporate a regularization term  into the optimization objective, which can be represented as:

{\fontsize{8pt}{8pt}\selectfont
\begin{equation} 
    \min_{\theta} \underset{(\boldsymbol{x}, {y}) \sim \mathcal{D}, \boldsymbol{x}'\in \mathcal{C}_{\varepsilon}(\boldsymbol{x})}{\mathbb{E}} [\text{CE}(f(\boldsymbol{x}), \boldsymbol{y}) - \beta \sum_{n_i \in \mathcal{N}} \ours(n_i, \boldsymbol{x}, \boldsymbol{x}')],
\end{equation}
}

\noindent
where $\beta$ is a hyperparameter that controls the importance of the regularization term, and $\mathcal{N}$ denotes the set of neurons in the network. 

Adapting concepts from adversarial training, the loss function can be reformulated to maximize the minimal (minimize the negative minimal) consistency of neural behavior across different inputs within the neighborhood domain:
{\fontsize{8pt}{8pt}\selectfont
\begin{equation} 
    \ourloss(\boldsymbol{x}, {y}) = \text{CE}(f(\boldsymbol{x}), \boldsymbol{y}) - \beta \underset{\boldsymbol{x}'\in \mathcal{C}_{\varepsilon}(\boldsymbol{x})}{\min} \sum_{n_i \in \mathcal{N}} \ours(n_i, \boldsymbol{x}, \boldsymbol{x}') .
\end{equation}
}

The consistency metric presented in Equation~\ref{eq: original_nbc} is a discrete measure and cannot be directly integrated into the loss function. Therefore, we employ a continuous metric to approximate the consistency of neural behavior across different inputs within the neighborhood domain, as shown in Algorithm~\ref{alg: NBC}. This algorithm calculates the NBC for the neural network $f$ when fed with $\boldsymbol{x}$ and $\boldsymbol{x}'$.

\begin{algorithm}[tb]
    \caption{Calculation of NBC}
    \label{alg: NBC}
    \textbf{Input}: Neural network $f$, input $\boldsymbol{x}$, neighbor input $\boldsymbol{x}'$

    \textbf{Output}: Neural behavior consistency between $\boldsymbol{x}$ and $\boldsymbol{x}'$
    
    \begin{algorithmic}[1]
    \STATE $s\gets 0$
    
    \FOR{the $i$-th layer $l_i$ in hidden layers}
        \STATE $\boldsymbol{v} \gets [f^{(i)}(\boldsymbol{x})_1, f^{(i)}(\boldsymbol{x})_2, \dots, f^{(i)}(\boldsymbol{x})_{m[i]}]^\text{T}$
        \STATE $\boldsymbol{v}' \gets [f^{(i)}(\boldsymbol{x}')_1, f^{(i)}(\boldsymbol{x}')_2, \dots, f^{(i)}(\boldsymbol{x}')_{m[i]}]^\text{T}$
        \STATE $nbc \gets \frac{\boldsymbol{v} \cdot \boldsymbol{v}'}{|\boldsymbol{v}| \cdot |\boldsymbol{v}'|}$ \label{alge: NBC: nbc}
        \STATE $s \gets s + \frac{nbc}{\gamma[i]}$ \label{alge: NBC: s}
    \ENDFOR
    \STATE $s \gets s - \textsc{KL}(f(\boldsymbol{x}) || f(\boldsymbol{x}'))$
    \STATE \textbf{return} $s$
    \end{algorithmic}
\end{algorithm}

During the calculation of \ours, the NBC value $s$ is initially set to zero. Then 
iterating over the hidden layers of $f$, the algorithm assesses the consistency of each neuron between the original and adversarial images, incrementally updating the NBC. The final NBC is computed as the sum of the scaled neuron consistencies across all layers.

Due to the characteristics of gradient backpropagation, the layer close to the output layer has a greater impact on the network's behavior. Therefore, we use the KL divergence as a consistency metric for the output layer, serving as a regularization term to ensure that the network's output remains consistent across different inputs. This can be calculated as:
{\fontsize{8pt}{8pt}\selectfont
\begin{equation} 
    \textsc{KL}(f(\boldsymbol{x}) || f(\boldsymbol{x}')) = \sum_{i} f(\boldsymbol{x})_i \log \frac{f(\boldsymbol{x})_i}{f(\boldsymbol{x}')_i}.
\end{equation}
}

For the hidden layers, to prevent gradient explosion or vanishing, we use cosine similarity as a continuous metric to approximate the consistency of behavior. Cosine similarity outputs a value in the range of [0,1], making it more suitable for training neural networks with high-dimensional intermediate layers.

To balance the impact of layers with different numbers of neurons, we scale the NBC value by the factor $\gamma[i]$ at line~\ref{alge: NBC: s} of Algorithm~\ref{alg: NBC}. Our intuition is to prioritize layers with smaller dimensions, as their more consistent behavior can help propagate constraints through the network. Additionally, layers near the input and output often have fewer neurons. Applying constraints to these layers directly influences the forward and backward propagation processes, which can accelerate the convergence of the target loss. In contrast, over-constraining the middle layers, which contain more neurons and extract more complex features, could limit the model's expressive power.
More generally, applying stricter penalties to layers may decrease accuracy but increase the proportion of stable neurons. For this reason, we mainly impose constraints on layers with fewer neurons.
In our experiments, we apply the factor $\gamma[i]=2^{r[i]}$ to balance the impact of layers with different numbers of neurons, where $r[i]$ means the number of neurons in the $i$-th hidden layer $m[i]$ is the $r[i]$-th smallest number in all the numbers of neurons in hidden layers.

\begin{algorithm}[t]
    \caption{Calculation of $\ourloss$}
    \label{alg:loss_algorithm}
    \textbf{Input}: Neural network $f$, input $\boldsymbol{x}$, label $\boldsymbol{y}$, perturbation $\varepsilon$, number of perturbation steps $k$, step size $\alpha$, NBC regularization hyperparameter $\beta$\\
    \textbf{Output}: Final \ours loss $\ourloss$
    \begin{algorithmic}[1] 
        \STATE Generate a random starting point $\boldsymbol{x}' \in \mathcal{C}_\epsilon(\boldsymbol{x})$
        \FOR{$i$ from $1$ to $k$}
            \STATE $\Delta \boldsymbol{x} \gets \frac{\partial \ours(f, \boldsymbol{x}, \boldsymbol{x}')}{\partial \boldsymbol{x}'}$
            \STATE $\boldsymbol{x}' \gets \boldsymbol{x}' - \alpha \Delta \boldsymbol{x}$
            \STATE $\boldsymbol{x}' \gets \text{clip}(\boldsymbol{x}', \boldsymbol{x}, \varepsilon)$
        \ENDFOR
        \STATE $\ourloss \gets \text{CE}(f(\boldsymbol{x}), \boldsymbol{y}) - \beta \cdot \ours(f, \boldsymbol{x}, \boldsymbol{x}')$
        \STATE \textbf{return} $\ourloss$
    \end{algorithmic}
\end{algorithm}

Algorithm~\ref{alg:loss_algorithm} outlines the \ours loss calculation process. This algorithm uses the idea of adversarial training to find an adversarial input $\boldsymbol{x}'$ that minimizes the NBC loss between the input $\boldsymbol{x}$ and $\boldsymbol{x}'$. 
The algorithm starts by selecting a random adversarial input $\boldsymbol{x}'$ within the $\varepsilon$-neighborhood of the original input $\boldsymbol{x}$. The adversarial input is then iteratively perturbed over $k$ steps, 
with each perturbation adjusting the adversarial input based on the gradients of the NBC concerning $\boldsymbol{x}'$
, scaled by the step size $\alpha$. The adversarial input is clipped after each perturbation step to ensure it remains within the perturbation range allowed by the original input. The final \ours loss is computed as the summation of the cross-entropy loss and the negation of scaled NBC (calculated according to Algorithm~\ref{alg: NBC}) between the original input and the adversarial input. A detailed discussion of the hyperparameters $\gamma$ in Algorithm~\ref{alg: NBC} and $\beta$ in Algorithm~\ref{alg:loss_algorithm} is provided in supplementary material~\cite{liu2024trainingverificationfriendlyneuralnetworks}.

We train the network to maximize the consistency of neuron behavior across varied inputs within the neighborhood, thereby reducing the number of unstable neurons and tightening neuron bounds. As a result, the network trained with the \ours loss is friendly to formal verification methods, as it exhibits fewer unstable neurons and more precise bounds.

\section{Evaluation}\label{sec: evaluation}

In this section, we evaluate our method to address the following research questions:

\noindent\textbf{RQ1:}
Can networks trained with our method maintain verification-friendly properties across various network architectures and perturbation radii?

\noindent\textbf{RQ2:}
Can our method be effectively integrated with existing training methods?

\noindent\textbf{RQ3:}
How does the performance of our method compare to existing methods when their accuracies are close?

\begin{figure*}[t]
    \centering
    \includegraphics[page=2]{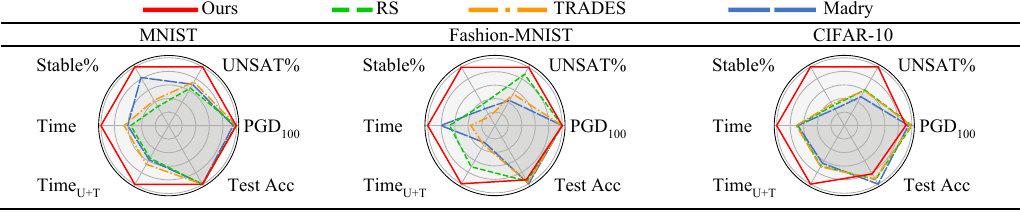}
    \caption{Overview of last epoch results. For each metric at each perturbation radius, the best-performing value is defined as 100, with other values scaled proportionally. We average all the scaled values of each model at each perturbation radius in each dataset to obtain the final score. For each metric, the larger value indicates better performance.}
    \label{fig: overview_total}
\end{figure*}

\begin{table*}[t]
    \centering
    \begin{tabular}{c|l|cccc|cccc|cccc}
    \toprule
    \multicolumn{2}{c|}{Model} & \multicolumn{4}{c|}{M1} & \multicolumn{4}{c|}{M2} & \multicolumn{4}{c}{M3} \\ \hline
    \multicolumn{2}{c|}{Method} & T & M & R & O & T & M & R & O & T & M & R & O \\ \hline
    \multicolumn{2}{c|}{Test Acc.} & \textbf{99.0}& 98.2 & 98.9 & 98.8 & 99.1 & 98.3 & \textbf{99.2}& 98.9 & 99.1 & 99.3 & \textbf{99.3}& 99.0 \\ \hline
    \multirow{5}{*}{$\varepsilon$=0.1} & UNSAT\% & 96.6 & 95.0 & 96.4 & \textbf{97.2} & 96.8 & 93.0 & 96.9 & \textbf{97.0} & 94.5 & 87.8 & 86.0 & \textbf{96.6} \\
     & Stable\% & 72.0 & 84.7 & 68.0 & \textbf{84.7} & 69.4 & 81.5 & 65.9 & \textbf{86.6} & 78.4 & 69.3 & 48.4 & \textbf{92.1} \\
     & Time (s) & 3.6 & 4.7 & 4.0 & \textbf{3.0} & 3.7 & 5.2 & 3.6 & \textbf{3.2} & 9.2 & 19.0 & 20.6 & \textbf{4.0} \\
     & Time$_\text{U+T}$ (s) & 3.5 & 4.5 & 3.9 & \textbf{2.9} & 3.6 & 5.0 & 3.5 & \textbf{3.1} & 8.9 & 19.1 & 20.7 & \textbf{3.9} \\
     & $\text{PGD}_{100}$ & 97.6 & 95.9 & \textbf{97.7} & \textbf{97.7} & 97.8 & 94.3 & \textbf{97.9} & 97.9 & 98.0 & 97.6 & \textbf{98.2} & 97.9 \\
    \hline
    \multirow{5}{*}{$\varepsilon$=0.2} & UNSAT\% & 89.5 & 86.7 & 79.0 & \textbf{92.2} & 85.6 & 71.3 & 71.8 & \textbf{91.8} & 2.3 & 0.2 & 1.5 & \textbf{46.3} \\
     & Stable\% & 34.8 & \textbf{65.7} & 25.2 & 64.5 & 27.7 & 66.1 & 20.5 & \textbf{68.0} & 17.2 & 10.0 & 4.3 & \textbf{68.7} \\
     & Time (s) & 13.7 & 12.8 & 28.0 & \textbf{7.3} & 20.8 & 30.2 & 39.1 & \textbf{8.7} & 111.8 & 113.0 & 113.1 & \textbf{68.4} \\
     & Time$_\text{U+T}$ (s) & 14.1 & 13.1 & 29.7 & \textbf{7.3} & 21.6 & 33.8 & 41.2 & \textbf{8.8} & 117.5 & 119.8 & 118.3 & \textbf{71.9} \\
     & $\text{PGD}_{100}$ & \textbf{95.7} & 91.8 & 95.0 & 95.6 & 96.0 & 88.8 & 95.8 & \textbf{96.4} & \textbf{96.1} & 95.5 & 96.0 & 96.0 \\
    \hline
    \multirow{5}{*}{$\varepsilon$=0.3} & UNSAT\% & 29.4 & 35.5 & 7.0 & \textbf{60.1} & 13.3 & 8.2 & 3.9 & \textbf{47.4} & \textbf{0.0} & \textbf{0.0} & \textbf{0.0} & \textbf{0.0} \\
     & Stable\% & 6.2 & 40.7 & 2.4 & \textbf{47.8} & 2.9 & 37.2 & 1.4 & \textbf{53.9} & 0.1 & 0.0 & 0.0 & \textbf{52.5} \\
     & Time (s) & 81.1 & 64.3 & 94.3 & \textbf{47.8} & 95.4 & 84.5 & 99.8 & \textbf{63.5} & 108.4 & \textbf{106.6} & 108.3 & 107.7 \\
     & Time$_\text{U+T}$ (s) & 91.5 & 80.0 & 113.8 & \textbf{55.0} & 108.5 & 113.0 & 116.6 & \textbf{72.9} & \textbf{120.0} & \textbf{120.0} & \textbf{120.0} & \textbf{120.0} \\
     & $\text{PGD}_{100}$ & \textbf{93.1} & 83.3 & 89.3 & 91.3 & \textbf{92.8} & 80.0 & 91.8 & 91.4 & \textbf{92.2} & 91.3 & 91.6 & 91.9 \\
    \bottomrule
    \end{tabular}
          \caption{Networks trained with $\varepsilon=0.3$ on MNIST datasets. The best results are highlighted in bold. Verified under $\varepsilon$=0.1, 0.2, 0.3. T:TRADES, M:Madry, R:ReLU Stable, O:Ours.}
    \label{tab: mnist_results}
\end{table*}

\begin{table*}[t]
    \centering\small
    \begin{tabular}{c|l|ccc|ccc|ccc}
    \toprule
    \multicolumn{2}{c|}{Model} & \multicolumn{3}{c|}{M1} & \multicolumn{3}{c|}{M2} & \multicolumn{3}{c}{M3} \\ \hline 
    \multicolumn{2}{c|}{Method} & T* & M* & R* & T* & M* & R* & T* & M* & R* \\ \hline 
    \multicolumn{2}{c|}{Test Acc.} & -0.1 & -3.9 & -3.5 & -0.1 & -5.7 & -2.8 & -0.1 & -0.4 & -0.5 \\ \hline 
    \multirow{5}{*}{$\varepsilon$=0.1} & UNSAT\% & \textbf{+0.6} & -7.6 & -4.3 & -0.2 & -11.2 & -3.8 & \textbf{+2.0} & \textbf{+9.0} & \textbf{+10.6} \\ 
     & Stable\% & \textbf{+12.7} & \textbf{+12.8} & \textbf{+24.4} & \textbf{+16.4} & \textbf{+15.8} & \textbf{+25.6} & \textbf{+13.7} & \textbf{+25.3} & \textbf{+44.4} \\ 
     & Time (s) & \textbf{-0.7} & \textbf{-1.1} & \textbf{-0.7} & +1.1 & +0.9 & \textbf{-0.3} & \textbf{-3.4} & \textbf{-14.8} & \textbf{-16.3} \\ 
     & Time$_\text{U+T}$ (s) & \textbf{-0.6} & \textbf{-1.4} & \textbf{-1.0} & +1.1 & +0.8 & \textbf{-0.5} & \textbf{-3.4} & \textbf{-15.1} & \textbf{-16.7} \\ 
     & $\text{PGD}_{100}$ & \textbf{+0.1} & -6.1 & -4.3 & -4.0 & -7.9 & -3.5 & -0.2 & \textbf{+0.3} & -0.4 \\ 
    \hline 
    \multirow{5}{*}{$\varepsilon$=0.2} & UNSAT\% & \textbf{+2.7} & -10.0 & \textbf{+9.4} & \textbf{+4.9} & -11.8 & \textbf{+18.0} & \textbf{+43.0} & \textbf{+83.2} & \textbf{+87.0} \\ 
     & Stable\% & \textbf{+29.7} & \textbf{+28.1} & \textbf{+57.9} & \textbf{+37.9} & \textbf{+26.2} & \textbf{+60.1} & \textbf{+51.0} & \textbf{+70.6} & \textbf{+74.6} \\ 
     & Time (s) & \textbf{-6.4} & \textbf{-6.8} & \textbf{-23.9} & \textbf{-7.2} & \textbf{-12.3} & \textbf{-34.7} & \textbf{-41.9} & \textbf{-88.0} & \textbf{-97.3} \\ 
     & Time$_\text{U+T}$ (s) & \textbf{-6.8} & \textbf{-7.8} & \textbf{-26.0} & \textbf{-7.7} & \textbf{-11.5} & \textbf{-37.2} & \textbf{-44.4} & \textbf{-94.4} & \textbf{-102.5} \\ 
     & $\text{PGD}_{100}$ & -0.2 & -9.9 & -4.2 & -2.2 & -12.8 & -4.1 & -0.2 & \textbf{+0.4} & -0.3 \\ 
    \hline 
    \multirow{5}{*}{$\varepsilon$=0.3} & UNSAT\% & \textbf{+30.7} & \textbf{+16.8} & \textbf{+72.4} & \textbf{+30.4} & -0.6 & \textbf{+75.1} & 0.0 & \textbf{+4.4} & \textbf{+38.9} \\ 
     & Stable\% & \textbf{+41.6} & \textbf{+46.2} & \textbf{+69.1} & \textbf{+47.7} & \textbf{+39.5} & \textbf{+64.9} & \textbf{+51.7} & \textbf{+57.4} & \textbf{+53.8} \\ 
     & Time (s) & \textbf{-33.3} & \textbf{-44.1} & \textbf{-86.2} & \textbf{-27.2} & \textbf{-32.6} & \textbf{-90.6} & \textbf{-0.5} & \textbf{-3.9} & \textbf{-40.8} \\ 
     & Time$_\text{U+T}$ (s) & \textbf{-36.5} & \textbf{-52.0} & \textbf{-105.9} & \textbf{-31.2} & \textbf{-6.4} & \textbf{-107.4} & 0.0 & \textbf{-4.4} & \textbf{-44.7} \\ 
     & $\text{PGD}_{100}$ & -1.5 & -15.2 & -3.0 & \textbf{+1.1} & -21.6 & -4.3 & \textbf{+0.2} & \textbf{+1.0} & -0.4 \\ 
    \bottomrule
    \end{tabular}   
    \caption{Networks trained with each method combined with our method on the MNIST dataset with $\varepsilon=0.3$. Improved results are highlighted in bold. T*: TRADES+Ours, M*: Madry+Ours, R*: ReLU Stable+Ours.}
    \label{tab: mnist_combine}
\end{table*}

\subsection{Experimental Setup}

Experiments are conducted on a server with 128 Intel Xeon Platinum 8336C CPUs, 128GB memory, and four NVIDIA GeForce RTX 4090 GPUs, running Debian GNU/Linux 10 (Buster). We use Python 3.11.7 and PyTorch 2.1.2 for implementation. Other settings are as follows.

\noindent
\textbf{Dataset}. 
Networks are trained on three widely used datasets: MNIST~\cite{MNIST}, Fashion-MNIST~\cite{fashion_mnist}, and CIFAR-10~\cite{cifar10}.

\noindent
\textbf{Network Architecture}. 
We select neural networks of varying sizes from VNN-COMP \cite{VNN2021,VNN2022} to assess the effectiveness of the methods used. For the MNIST and Fashion-MNIST datasets, we chose the M1 (cnn\_4\_layer), M2 (relu\_stable), and M3 (conv\_big) models, with approximately 0.16M, 0.17M, and 1.9M parameters, respectively. For the CIFAR-10 dataset, we select the C1 (marabou\_medium), C2 (marabou\_large), and C3 (conv\_big) models, containing about 0.17M, 0.34M, and 2.4M parameters, respectively. Network architectures are detailed in supplementary material~\cite{liu2024trainingverificationfriendlyneuralnetworks}.

\noindent
\textbf{Baselines}.
We choose Relu Stable~\cite{relustable2019} as a baseline, as it shares the most similarities with our approach.
TRADES~\cite{trades} and Madry~\cite{DBLP:MadryMSTV18}, two commonly used adversarial training methods, are selected to show that directly using our method can at least achieve robustness comparable to classical robust training methods and combining our method with existing methods can improve verification performance.

\noindent
\textbf{Training}. 
The batch size is set to 128, and using the Adam optimizer. For RQ1, networks are trained under default settings for 400 epochs. For RQ2, each network is trained for 200 epochs using the original method, followed by an additional 200 epochs combining the original method with our approach, or vice versa. For RQ3, we train a base model using the CE loss, then fine-tune it separately using RS, Madry, TRADES, and our method, ensuring that each method maintains accuracy within a specified range.

\noindent
\textbf{Verification}.
We use $\alpha,\beta$-CROWN, a state-of-the-art verification tool that performs the best in VNN-COMP competitions, to verify the properties of the trained networks. For each dataset, we select $k$ images from each of the 10 categories in the test set. For each image $\boldsymbol{x}$ and its ground truth label $y$, we verify the property that the network's output label remains $y$ for input $\boldsymbol{x}$ under each perturbation $\varepsilon$. We set $k=100$ and a timeout of 120 seconds for MNIST and Fashion-MNIST, and $k=20$ with a timeout of 180 seconds for CIFAR-10.

\smallskip\noindent
\textbf{Metrics}. The metrics used in our evaluation are as follows:

\begin{itemize}
    \item UNSAT\%: The percentage of properties verified to hold (UNSAT), indicating the network's overall verification effectiveness and robustness.
    \item Stable\%: The average percentage of stable neurons, calculated as: $\frac{\sum_{1}^{N}s_i}{N}$, where $s_i$ is stable neuron ratio of the $i$-th property, and $N$ is the total number of properties.
    \item Time: The average time required to verify a property.
    \item Time$_\text{U+T}$: The average time is taken to verify properties that result in UNSAT or Timeout. This metric is more indicative of the efficiency of the verification process, as it excludes the time taken to verify properties that are SAT, which can be verified quickly by attacking the network. 
    \item $\text{PGD}_{100}$: The network's accuracy under a PGD attack for 100 steps with a perturbation $\varepsilon$.
\end{itemize}

\subsection{Results of the Last Epoch}
Figure~\ref{fig: overview_total} shows the overall evaluation results of networks trained on MNIST, Fashion-MNIST, and CIFAR-10. Since some metric values are too large or small, we set the best-performing value to 100, with other values scaled proportionally for better visualization.
For the metrics Time and Time$_\text{U+T}$, we use the reciprocal of the values to make the visualization more intuitive. Therefore, the larger the metric value in the table, the better the performance.

We observe that our method outperforms others in terms of verification time, stable neuron ratio, and verified ratio (UNSAT\%). While our method slightly lags in accuracy and PGD accuracy, it generally maintains comparable accuracy to the other methods. This result represents a comprehensive evaluation across various perturbation radii and models, indicating that networks of various architectures trained using our method are more verification-friendly at different perturbation radii than those trained using other methods.

Table~\ref{tab: mnist_results} shows the detailed results for networks trained on MNIST. Detailed results for Fashion-MNIST and CIFAR-10 are provided in the supplementary material. As shown in Table~\ref{tab: mnist_results}, traditional adversarial training methods enhance network robustness, yet verifiability declines as the perturbation radius increases. This suggests the need for new training techniques that support effective verification.

Regarding UNSAT\%, our method consistently outperforms others, especially at higher perturbation radii. For instance, when $\varepsilon=0.2$, our verified ratio reached 46.3\%, more than 20 times that of the next best-performing method and over 231 times that of the least effective method on the M3 model. Unlike other methods, where UNSAT\% significantly drops as the radius increases, our method maintains a high verified ratio across all tested radii. 

As for stable neurons, our method excels across nearly all models and perturbation settings. Notably, in the M3 model at $\varepsilon=0.3$, while competing methods exhibited virtually no stable neurons, our method preserved a stable neuron ratio above 50\%. A higher proportion of stable neurons means that if the verification time is increased, our method is more likely to be successfully verified (as the search space is significantly reduced).

Regarding average verification time, including UNSAT and Timeout issues, our method generally requires less time across all verification tasks. An exception occurs in the M3 model at a 0.3 perturbation radius, where verification failed to confirm all properties of our model. In contrast, the model trained using Madry quickly verified many properties as violated, thus shortening the verification times. Furthermore, under a 100-step PGD attack, our method achieved comparable accuracy to these adversarial training approaches.
These results also hold for the Fashion-MNIST and CIFAR-10 datasets (see supplementary material).

\begin{custombox}
\textbf{Answer to RQ1}: 
Our method maintains high stable neuron ratios, UNSAT\%, and robustness across various perturbation radii. Overall, networks trained with our method preserve their verification-friendly properties across different network architectures and radii.
\end{custombox}

\subsection{Combination with Existing Methods}

Table~\ref{tab: mnist_combine} shows the results of networks trained with our method combined with other methods on MNIST dataset.
All the results show that our method combined with existing methods improves the ratio of stable neurons. In most cases, our method combined with existing methods increases the ratio of verified properties (UNSAT\%) and reduces the time required for verification greatly.

It is worth noting that when our method is combined with a baseline method, the performance is particularly outstanding in larger network structures. For example, in the M3 model. Our method combined with the Madry and RS methods respectively increased the verified ratio by 83.2\% and 87.0\% at $\varepsilon=0.2$, while also reducing verification time by 94.4 seconds and 102.5 seconds. 
This means that our method makes it possible to verify larger network structures. The same results are also reflected in the Fashion-MNIST and CIFAR-10 datasets provided in the supplementary material.

Combining our method with existing methods sacrifices some accuracy and adversarial accuracy but leads to improved verification and neural stability. The improvement is particularly significant for larger models, meaning that our method enables the verification of larger network structures using existing verification tools. Similar results are also observed in the Fashion-MNIST and CIFAR-10 datasets, as shown in the supplementary material.

\begin{custombox}
\textbf{Answer to RQ2}: Our method, when combined with other methods, improves the verification-friendly properties of the network. However, a trade-off between accuracy and verifiability is required.
\end{custombox}

\subsection{Comparison under Close Accuracy}

\begin{figure}[htbp]
    \centering
    \includegraphics[page=3]{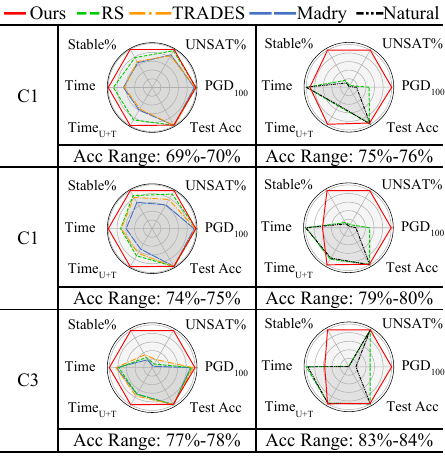}
    \caption{Networks trained with $\varepsilon={2}/{255}$ on CIFAR-10 dataset and evaluated on $\varepsilon={2}/{255}$. Methods unable to achieve a given accuracy range are omitted. 'Natural' means the network is trained with only cross-entropy loss.}
    \label{fig: cifar_results2}
\end{figure}

The accuracy of networks trained with default parameters varied significantly on the CIFAR-10 dataset.
To compare the performance of networks with similar accuracy, we fine-tuned the parameters to ensure that the accuracy of each method remained within a certain range.

Figure~\ref{fig: cifar_results2} shows the results of networks trained on CIFAR-10.
Larger perturbation radii are very challenging for current verification methods, so we chose a smaller perturbation radius to train and evaluate the networks.

For UNSAT\%, our method performs best across different network models and different accuracy ranges and generally requires less verification time for timeouts and UNSAT results than other methods, indicating that our networks are easier to verify. When the models reach higher accuracy, our method has a slightly longer average time, as other methods are more easily to find counterexamples using attack methods, while our method is less vulnerable to attacks. 

Our accuracy under PGD is generally higher than that of other methods, with a few exceptions where it is slightly lower than the best-performing method. Our method consistently maintains a higher ratio of stable neurons than other methods. Notably, in the C3 model, our method maintains a high stable neuron ratio of up to 80.1\% while preserving high verification accuracy. This indicates that the upper bound of the search space is significantly reduced.

\begin{custombox}
\textbf{Answer to RQ3}: Experiments show that our method is more verification-friendly than other methods at higher and close accuracy levels, as it consistently maintains a high ratio of stable neurons, greater robustness, and requires less verification time. 
\end{custombox}

\section{Related Work}  \label{sec: related_work}

Traditional adversarial methods such as TRADES~\cite{trades} and Madry~\cite{DBLP:MadryMSTV18}  use adversarial training to improve the robustness of neural networks. However, these methods primarily impose constraints on the network's output, while our method imposes neuron behavior constraints on all the network layers. Recent works, such as HYDRA\cite{sehwag2020hydra}, integrate the pruning process with adversarial training, using a criterion that considers adversarial robustness during the pruning decision. This leads to more compact yet robust models. Wu et al.~\cite{wu2020adversarial} also found that perturbing the weight can improve the robustness of the network.

Certified training~\cite{de2022ibp,mirman2018differentiable,jovanovic2022paradox,zhang2019towards,xu2020automatic} introduces interval bound propagation (IBP) into the training process to improve network robustness but suffers from long training times.

The ReLU Stable method~\cite{relustable2019} ensures that the upper and lower bounds of the neurons have the same sign, pushing them away from zero in the same direction. This method can also incorporate ternary loss to train a SAT-friendly network~\cite{narodytska2019search}. In contrast, our method emphasizes the stability of neurons before and after perturbation, focusing on the consistency of neuron behavior. Furthermore, RS loss requires additional calculations for each neuron's bounds, whereas our method does not require extra information, making it easier to implement.

Linearity grafting~\cite{chen2022linearity} replaces unstable neurons with linear neurons to improve the robustness of the network. However, this method modifies the network architecture, whereas our method preserves the architecture.

The MILP-based method~\cite{baninajjar2023verification} uses MILP to post-process the network and make it sparse, improving verification efficiency. However, this may be time-consuming.

Pruning-based methods~\cite{relustable2019,training_for_verification} heuristically prune inactive or unstable neurons with little impact on the network's performance. The bias shaping method~\cite{training_for_verification} changes the bias of unstable neurons during training to stabilize them. These methods are either post-processing techniques or training tricks that can be incorporated with other methods to improve the verifiability of networks further.

\section{Conclusion} \label{sec: conclusion}

In this work, we propose a novel training method to develop a verification-friendly neural network by preserving neuron behavior consistency.
Our experiments on the MNIST, Fashion-MNIST, and CIFAR-10 datasets demonstrate that our method consistently enhances the network's verification-friendliness, as evidenced by a higher stable neuron ratio, comparable robustness, and faster verification speed across different perturbation radii.
Additionally, when combined with existing methods, our method shows improved verification efficiency. Our method also accelerates the verification process while maintaining model accuracy, which is a result that is rarely achieved by existing methods. 
In future work, we plan to explore the application of our method to more complex network architectures and datasets, as well as explain the underlying mathematical machinery of our method in more detail.

\newpage
\section*{Acknowledgements}
This work is partly supported by CAS Project for Young Scientists in Basic Research, Grant No.YSBR-040, ISCAS New Cultivation Project ISCAS-PYFX-202201, ISCAS Basic Research ISCAS-JCZD-202302 and the Ministry of Education, Singapore under its Academic Research Fund Tier 3 (Award ID: MOET32020-0004).

\bibliography{aaai25}
\appendix
\section{Detailed Experimental Setup}

\noindent
\textbf{Dataset Selection}.

We select three datasets: Fashion-MNIST, MNIST, and CIFAR-10 datasets for our experiments. The MNIST and CIFAR-10 datasets are widely used in early adversarial training research~\cite{DBLP:MadryMSTV18, trades} and similar studies~\cite{relustable2019, training_for_verification}. To demonstrate the generality of our method, we additionally included the Fashion-MNIST dataset. Fashion-MNIST is also a dataset for a ten-class classification task like the MNIST dataset, but the images in the Fashion-MNIST dataset are more complex and challenging compared to the MNIST dataset.

\noindent
\textbf{Network Architecture Selection}.

We select the following network architectures based on the network sizes used in previous works~\cite{relustable2019, training_for_verification}, the verification capabilities of neural network verification tools, and the generality of the networks. M1 and C1 are smaller networks, M2 and C2 are medium-sized networks, and M3 and C3 are larger networks. Compared to previous works, we mostly use convolutional networks instead of fully connected networks. This is because the use of convolutional networks is more common in real applications.
Moreover, M3 and C3 are even larger than the largest network structures used in similar works~\cite{relustable2019, training_for_verification} and are very close to the limits of neural network verification tools. Table~\ref{tab: network_architecture} provides detailed descriptions of the network architectures used in our experiments.

\noindent
\textbf{Training Details}.

For the MNIST and the Fashion-MNIST datasets, we directly use the original pictures to train each network. For the CIFAR-10 dataset, we randomly crop images to 32x32 pixels and augment the data by flipping images horizontally with a probability of 0.5. We also add padding of 4 pixels on each side before cropping.

We use the Adam optimizer to train all networks with a batch size of 128. The learning rate is set to 0.0001 for the MNIST and Fashion-MNIST datasets and 0.00001 for the CIFAR-10 dataset. We use the same number of training iterations for all datasets, which is 400 iterations. The random seed is set to 0 for reproducibility. All PGD-like adversarial training process uses the same hyperparameters as 10 steps and a step size of $\varepsilon/10$.

In the last epoch experiment, for the RS \cite{relustable2019}, Madry \cite{DBLP:MadryMSTV18}, and TRADES \cite{trades} methods, we used the same hyperparameters as in the original papers. 
As for our model, we use the hyperparameters $\beta=1$ for M1, M2 and M3 models, and $\beta=2, 5, 3$ for C1, C2, and C3 models, respectively. We select the hyperparameters based on the performance of the model on the validation set (generated by the original training set).

In the combination experiment, we find that different methods have different performances when using different combination orders. Based on the performance of the model on the validation set. For the RS method, we first train 200 epochs using the original RS loss and then combine it with our NBC loss for another 200 epochs. For the Madry method, we use the method combined with NBC loss for 200 epochs and then use the original Madry loss for another 200 epochs. For the TRADES method, we use the method combined with the NBC loss to train directly for 400 epochs. The parameters are the same as the original training process.

The application of these methods results mainly in a decrease in accuracy compared to direct training without these methods, which may be unacceptable in practical applications. Therefore, our objective is to evaluate the effectiveness of these methods while maintaining relatively high accuracy. We specified two sets of accuracy ranges for each model: one where the accuracy difference from the 'Natural' method is within 1\%-2\%, and another where the difference is within 6\%-7\%. We believe this setup effectively evaluates the effectiveness of each method under different accuracy requirements. 
Besides, as shown in Table~\ref{tab: mnist_results} and \ref{tab: cifar_results}, various metrics drop as the robustness radius increases, making the differences in metrics between various methods less apparent. The use of $\varepsilon=2/255$ to train the model and the same $\varepsilon$ to verify the model reveals the differences more clearly and better illustrates the issue.
In this experiment, we initially trained a baseline network using only cross-entropy loss (referred to as 'Natural' in the tables and figures). We then used this model as a pre-trained baseline and applied various methods to fine-tune it separately, aiming to achieve a relatively high accuracy.

\noindent
\textbf{Verification Tasks}.

To demonstrate that our methods have generalization capabilities, we train on the training set and generate local robustness verification tasks under different perturbation radii on the test set.  We use the same verification tasks for all methods to ensure fairness. 
\begin{table*}[t]
    \centering\small
    \begin{tabular}{c|c|c|c}
        \toprule
        Dataset & Architecture Name & \#Parameters & Network Architecture  \\
        \hline
        \multirow{12}{*}{\makecell[c]{(Fashion-)\\MNIST}} & \multirow{3}{*}{Cnn 4layer (M1)} & \multirow{3}{*}{166,406} & \mbox{Conv2d(1, 16, (4, 4), (2, 2), (1, 1))} \\
                                                                                    & & & \mbox{Conv2d(16, 32, (4, 4), (2, 2), (1, 1))} \\
                                                                                    & & &  \mbox{Linear(1568, 100), Linear(100, 10)}   \\  \cline{2-4}
                               & \multirow{3}{*}{Relu Stable (M2)} & \multirow{3}{*}{171,158} & \mbox{Conv2d(1, 16, (5, 5), (2, 2), (2, 2))} \\
                                                                                     & & & \mbox{Conv2d(16, 32, (5, 5), (2, 2), (2, 2))} \\
                                                                                     & & & \mbox{Linear(1568, 100), Linear(100, 10)} \\ \cline{2-4}
                               & \multirow{6}{*}{Conv Big (M3)} & \multirow{6}{*}{1,974,762} & \mbox{Conv2d(1, 32, (3, 3), (1, 1), (1, 1))} \\
                                                                                    & & & \mbox{Conv2d(32, 32, (4, 4), (2, 2), (1, 1))} \\
                                                                                    & & & \mbox{Conv2d(32, 64, (3, 3), (1, 1), (1, 1))} \\
                                                                                    & & &  \mbox{Conv2d(64, 64, (3, 3), (2, 2), (1, 1))} \\
                                                                                    & & &  \mbox{Linear(3316, 512), Linear(512, 512)}  \\
                                                                                    & & & \mbox{Linear(512, 10)} \\
        \hline
        \multirow{14}{*}{CIFAR-10} & \multirow{4}{*}{Marabou medium (C1)} & \multirow{4}{*}{165,498} &\mbox{Conv2d(3, 16, (4, 4), (2, 2), (2, 2))}\\
                                                                                           & & & \mbox{Conv2d(16, 32, (4, 4), (2, 2), (2, 2))}\\
                                                                                           & & & \mbox{Linear(1152, 128), Linear(128, 64)}\\
                                                                                           & & &  \mbox{Linear(64, 10)}  \\ \cline{2-4}
                                  & \multirow{4}{*}{Marabou large (C2)} & \multirow{4}{*}{338,346} & \mbox{Conv2d(3, 16, (4, 4), (2, 2), (2, 2))}\\
                                                                                          & & & \mbox{Conv2d(16, 32, (4, 4), (2, 2), (2, 2))}\\
                                                                                          & & & \mbox{Linear(2304, 128), Linear(128, 64)}\\
                                                                                          & & & \mbox{Linear(64, 10)}\\ \cline{2-4}
                                  & \multirow{6}{*}{Conv Big (C3)} & \multirow{6}{*}{2,466,858} &\mbox{Conv2d(3, 32, (3, 3), (1, 1), (1, 1))}\\
                                                                                         & & & \mbox{Conv2d(32, 32, (4, 4), (2, 2), (1, 1))}\\
                                                                                         & & & \mbox{Conv2d(32, 64, (3, 3), (1, 1), (1, 1))}\\
                                                                                         & & & \mbox{Conv2d(64, 64, (3, 3), (2, 2), (1, 1))}\\
                                                                                         & & & \mbox{Linear(4096, 512), Linear(512, 512)}\\
                                                                                         & & & \mbox{Linear(512, 10)}\\ 
        \bottomrule
    \end{tabular}
    \caption{Network Architectures for Experiments}
    \label{tab: network_architecture}
\end{table*}

\begin{table*}[!t]
    \centering\small
    \begin{tabular}{c|l|cccc|cccc|cccc}
    \toprule
    \multicolumn{2}{c|}{Model} & \multicolumn{4}{c|}{M1} & \multicolumn{4}{c|}{M2} & \multicolumn{4}{c}{M3} \\ \hline
    \multicolumn{2}{c|}{Method} & T & M & R & O & T & M & R & O & T & M & R & O \\ \hline
    \multicolumn{2}{c|}{Test Acc.} & 84.0 & \textbf{84.4}& 81.3 & 82.1 & 83.8 & \textbf{86.0}& 80.5 & 78.3 & 87.2 & \textbf{89.3}& 86.2 & 81.6 \\ \hline
    \multirow{5}{*}{$\varepsilon$=0.1} & UNSAT\% & 53.5 & 30.6 & \textbf{70.1} & 69.1 & 49.3 & 2.0 & \textbf{70.0} & 65.5 & 16.3 & 0.2 & 27.1 & \textbf{63.3} \\
     & Stable\% & 36.6 & 66.7 & 68.0 & \textbf{78.9} & 31.1 & 47.7 & 65.2 & \textbf{86.2} & 4.4 & 3.8 & 6.7 & \textbf{84.4} \\
     & Time (s) & 25.2 & 26.5 & 9.5 & \textbf{6.8} & 29.7 & 19.3 & 9.4 & \textbf{8.0} & 79.7 & 33.0 & 68.4 & \textbf{20.4} \\
     & Time$_\text{U+T}$ (s) & 32.7 & 46.5 & 9.2 & \textbf{6.3} & 40.5 & 103.5 & 9.9 & \textbf{7.9} & 97.8 & 118.6 & 82.7 & \textbf{22.1} \\
     & $\text{PGD}_{100}$ & 70.5 & 49.5 & \textbf{73.8} & 71.8 & 68.9 & 30.5 & \textbf{73.6} & 71.1 & 77.5 & 31.8 & \textbf{78.1} & 74.7 \\
    \hline
    \multirow{5}{*}{$\varepsilon$=0.2} & UNSAT\% & 24.8 & 6.0 & \textbf{55.4} & 54.5 & 9.9 & 0.1 & \textbf{54.2} & 46.0 & 0.0 & 0.0 & 0.8 & \textbf{42.4} \\
     & Stable\% & 15.1 & 34.0 & 37.7 & \textbf{63.3} & 9.8 & 17.3 & 34.9 & \textbf{70.1} & 0.0 & 0.2 & 0.1 & \textbf{63.2} \\
     & Time (s) & 48.6 & 42.7 & 15.7 & \textbf{9.5} & 61.2 & \textbf{13.8} & 16.9 & 22.0 & 82.0 & \textbf{32.1} & 83.6 & 36.9 \\
     & Time$_\text{U+T}$ (s) & 78.1 & 104.5 & 18.8 & \textbf{10.4} & 105.4 & 119.7 & \textbf{22.1} & 32.0 & 120.0 & 120.0 & 118.9 & \textbf{48.7} \\
     & $\text{PGD}_{100}$ & 62.7 & 42.5 & \textbf{66.3} & 61.8 & 62.6 & 32.5 & \textbf{67.1} & 65.0 & 68.8 & 31.2 & \textbf{70.8} & 67.9 \\
    \hline
    \multirow{5}{*}{$\varepsilon$=0.3} & UNSAT\% & 5.3 & 0.4 & 31.2 & \textbf{35.6} & 0.2 & 0.0 & \textbf{30.0} & 17.4 & 0.0 & 0.0 & 0.0 & \textbf{17.4} \\
     & Stable\% & 6.2 & 15.2 & 14.4 & \textbf{54.2} & 1.9 & 5.8 & 12.6 & \textbf{57.4} & 0.0 & 0.0 & 0.0 & \textbf{43.8} \\
     & Time (s) & 52.6 & 34.6 & 30.6 & \textbf{16.4} & 49.5 & \textbf{10.4} & 33.2 & 41.2 & 62.8 & \textbf{34.6} & 69.2 & 51.0 \\
     & Time$_\text{U+T}$ (s) & 109.9 & 118.5 & 50.7 & \textbf{27.0} & 119.6 & 120.0 & \textbf{56.8} & 79.8 & 120.0 & 120.0 & 120.0 & \textbf{85.4} \\
     & $\text{PGD}_{100}$ & 53.1 & 36.7 & \textbf{57.3} & 49.4 & 51.1 & 31.7 & 58.9 & \textbf{59.5} & 58.3 & 36.8 & \textbf{62.2} & 59.5 \\
    \bottomrule
    \end{tabular}
    \caption{Networks trained with $\varepsilon=0.3$ on Fashion-MNIST datasets. The best results are highlighted in bold. Verified under $\varepsilon$=0.1, 0.2, 0.3. T:TRADES, M:Madry, R:ReLU Stable, O:Ours.}
\label{tab: fashion_results}
\end{table*}

\begin{table*}[!t]
    \centering\small
    \begin{tabular}{c|l|cccc|cccc|cccc}
    \toprule
    \multicolumn{2}{c|}{Model} & \multicolumn{4}{c|}{C1} & \multicolumn{4}{c|}{C2} & \multicolumn{4}{c}{C3} \\ \hline
    \multicolumn{2}{c|}{Method} & T & M & R & O & T & M & R & O & T & M & R & O \\ \hline
    \multicolumn{2}{c|}{Test Acc.} & 59.2 & \textbf{62.8}& 56.0 & 51.8 & 60.9 & \textbf{69.5}& 61.8 & 54.9 & 69.5 & \textbf{75.5}& 74.1 & 64.9 \\ \hline
    \multirow{5}{*}{$\frac{2}{255}$} & UNSAT\% & 46.5 & \textbf{49.0} & 48.5 & 42.0 & 50.0 & \textbf{51.0} & 47.5 & 44.0 & 24.0 & 9.0 & 12.0 & \textbf{39.0} \\
     & Stable\% & 79.3 & 76.2 & 80.4 & \textbf{91.6} & 77.6 & 69.6 & 74.8 & \textbf{94.2} & 53.4 & 42.3 & 44.1 & \textbf{80.3} \\
     & Time (s) & 13.8 & 14.8 & 11.1 & \textbf{9.8} & 17.1 & 21.2 & 24.7 & \textbf{14.2} & 77.7 & 112.9 & 105.2 & \textbf{46.8} \\
     & Time$_\text{U+T}$ (s) & 16.2 & 18.0 & 11.1 & \textbf{7.7} & 20.6 & 27.5 & 33.9 & \textbf{15.8} & 114.9 & 158.1 & 150.9 & \textbf{68.3} \\
     & $\text{PGD}_{100}$ & 51.3 & \textbf{53.8} & 49.9 & 50.9 & 53.5 & \textbf{59.3} & 55.1 & 48.2 & 62.7 & 64.5 & \textbf{65.3} & 57.8 \\
    \hline
    \multirow{5}{*}{$\frac{4}{255}$} & UNSAT\% & 22.5 & 17.0 & 30.0 & \textbf{31.0} & 17.5 & 8.0 & 14.5 & \textbf{28.0} & 1.0 & 0.5 & 1.0 & \textbf{2.5} \\
     & Stable\% & 48.9 & 40.0 & 50.7 & \textbf{80.5} & 40.4 & 24.1 & 33.6 & \textbf{84.9} & 7.3 & 2.8 & 3.3 & \textbf{50.8} \\
     & Time (s) & 39.3 & 56.3 & 33.1 & \textbf{17.9} & 61.5 & 80.4 & 61.4 & \textbf{27.3} & 101.8 & 104.3 & 104.7 & \textbf{90.4} \\
     & Time$_\text{U+T}$ (s) & 85.6 & 118.8 & 65.0 & \textbf{29.3} & 121.5 & 156.5 & 127.6 & \textbf{51.6} & 176.9 & 178.7 & 177.0 & \textbf{172.1} \\
     & $\text{PGD}_{100}$ & 43.6 & \textbf{44.5} & 44.0 & 41.8 & 46.6 & \textbf{48.5} & 48.3 & 41.5 & 54.8 & 52.8 & \textbf{55.6} & 50.8 \\
    \hline
    \multirow{5}{*}{$\frac{6}{255}$} & UNSAT\% & 5.0 & 2.5 & 9.0 & \textbf{17.0} & 1.0 & 1.0 & 1.5 & \textbf{12.5} & 1.0 & 0.0 & 0.0 & \textbf{1.0} \\
     & Stable\% & 23.4 & 15.9 & 24.3 & \textbf{68.0} & 14.0 & 5.7 & 9.7 & \textbf{73.7} & 1.1 & 0.4 & 0.4 & \textbf{44.3} \\
     & Time (s) & 49.0 & 56.5 & 54.9 & \textbf{33.4} & 69.3 & 62.6 & 61.7 & \textbf{39.9} & 89.6 & 85.4 & 82.4 & \textbf{78.6} \\
     & Time$_\text{U+T}$ (s) & 153.2 & 167.9 & 140.8 & \textbf{87.8} & 175.2 & 175.0 & 172.6 & \textbf{112.3} & \textbf{176.3} & 180.0 & 180.0 & 178.6 \\
     & $\text{PGD}_{100}$ & 36.8 & 35.5 & \textbf{38.5} & 33.0 & 39.5 & 37.9 & \textbf{41.1} & 36.0 & \textbf{47.4} & 40.9 & 45.4 & 44.4 \\
    \hline
    \multirow{5}{*}{$\frac{8}{255}$} & UNSAT\% & 1.0 & 0.5 & 1.0 & \textbf{7.5} & 0.5 & 0.5 & 0.5 & \textbf{2.5} & 0.0 & 0.0 & 0.0 & \textbf{0.5} \\
     & Stable\% & 10.1 & 5.8 & 10.8 & \textbf{57.7} & 4.2 & 1.2 & 2.5 & \textbf{65.3} & 0.2 & 0.1 & 0.0 & \textbf{43.3} \\
     & Time (s) & 45.8 & 44.0 & 54.3 & \textbf{34.3} & 54.4 & \textbf{43.4} & 53.3 & 44.9 & 75.0 & \textbf{62.7} & 65.2 & 66.5 \\
     & Time$_\text{U+T}$ (s) & 172.7 & 176.0 & 174.3 & \textbf{124.3} & 176.8 & 179.6 & 176.6 & \textbf{163.4} & 180.0 & 180.0 & 180.0 & \textbf{177.8} \\
     & $\text{PGD}_{100}$ & 29.9 & 27.4 & \textbf{32.9} & 25.3 & 32.7 & 28.3 & \textbf{33.9} & 30.2 & \textbf{39.8} & 30.5 & 35.3 & 38.0 \\
    \bottomrule
    \end{tabular}
    \caption{Networks trained with $\varepsilon=8/255$ on CIFAR-10 dataset with default parameters. 
    The best results are highlighted in bold. 
    Verified under $\varepsilon \in \{ 2/255, 4/255, 6/255, 8/255\}$. T: TRADES, M: Madry, R: ReLU stable, O: Our method.}
\label{tab: cifar_results}
\end{table*}


\begin{table*}[!t]
    \centering\small
    \begin{tabular}{c|l|ccc|ccc|ccc}
    \toprule
    \multicolumn{2}{c|}{Model} & \multicolumn{3}{c|}{M1} & \multicolumn{3}{c|}{M2} & \multicolumn{3}{c}{M3} \\ \hline
    \multicolumn{2}{c|}{Method} & T* & M* & R* & T* & M* & R* & T* & M* & R* \\ \hline
    \multicolumn{2}{c|}{Test Acc.} & -7.7\% & -8.8\% & -2.9\% & -3.6\% & -10.0\% & -3.7\% & -5.7\% & -13.0\% & -7.5\% \\ \hline
    \multirow{5}{*}{$\varepsilon$=0.1} & UNSAT\% & \textbf{+1.9} & \textbf{+6.0} & -8.3 & -2.6 & \textbf{+28.0} & -6.4 & \textbf{+19.5} & \textbf{+48.5} & \textbf{+37.6} \\
     & Stable\% & \textbf{+52.7} & \textbf{+17.2} & \textbf{+14.8} & \textbf{+53.2} & \textbf{+28.9} & \textbf{+21.4} & \textbf{+71.6} & \textbf{+73.9} & \textbf{+82.6} \\
     & Time (s) & \textbf{-17.6} & \textbf{-5.0} & \textbf{-3.6} & \textbf{-14.4} & +8.5 & \textbf{-4.2} & \textbf{-32.7} & \textbf{-5.2} & \textbf{-57.1} \\
     & Time$_\text{U+T}$ (s) & \textbf{-25.1} & \textbf{-10.0} & \textbf{-4.3} & \textbf{-18.7} & \textbf{-52.7} & \textbf{-6.1} & \textbf{-34.6} & \textbf{-82.3} & \textbf{-72.7} \\
     & $\text{PGD}_{100}$ & -11.1 & \textbf{+7.6} & -9.5 & -11.4 & \textbf{+27.6} & -7.9 & -9.6 & \textbf{+36.5} & -10.7 \\
    \hline
    \multirow{5}{*}{$\varepsilon$=0.2} & UNSAT\% & \textbf{+11.9} & -4.5 & -7.7 & -3.4 & \textbf{+0.5} & -2.7 & \textbf{+0.1} & \textbf{+0.5} & \textbf{+49.1} \\
     & Stable\% & \textbf{+62.1} & \textbf{+29.5} & \textbf{+28.8} & \textbf{+54.0} & \textbf{+42.7} & \textbf{+38.5} & \textbf{+47.2} & \textbf{+45.8} & \textbf{+76.4} \\
     & Time (s) & \textbf{-34.5} & \textbf{-19.1} & \textbf{-8.0} & \textbf{-20.7} & +14.4 & \textbf{-9.9} & \textbf{-11.4} & +37.6 & \textbf{-66.6} \\
     & Time$_\text{U+T}$ (s) & \textbf{-55.6} & +7.3 & \textbf{-11.4} & \textbf{-1.8} & \textbf{-2.4} & \textbf{-15.9} & \textbf{-0.1} & \textbf{-0.6} & \textbf{-101.1} \\
     & $\text{PGD}_{100}$ & -16.8 & -8.1 & -14.7 & -21.1 & \textbf{+2.7} & -12.0 & -11.0 & \textbf{+29.6} & -13.8 \\
    \hline
    \multirow{5}{*}{$\varepsilon$=0.3} & UNSAT\% & \textbf{+11.7} & -0.4 & -2.4 & -0.1 & 0.0 & \textbf{+5.5} & 0.0 & 0.0 & \textbf{+30.6} \\
     & Stable\% & \textbf{+58.0} & \textbf{+40.0} & \textbf{+36.5} & \textbf{+49.0} & \textbf{+48.6} & \textbf{+47.8} & \textbf{+17.8} & \textbf{+2.1} & \textbf{+63.6} \\
     & Time (s) & \textbf{-30.2} & \textbf{-25.6} & \textbf{-17.4} & \textbf{-29.7} & \textbf{-1.6} & \textbf{-22.5} & \textbf{-11.2} & \textbf{-16.5} & \textbf{-43.8} \\
     & Time$_\text{U+T}$ (s) & \textbf{-50.5} & +1.5 & \textbf{-27.1} & \textbf{-0.3} & 0.0 & \textbf{-42.2} & 0.0 & \textbf{-120.0} & \textbf{-80.8} \\
     & $\text{PGD}_{100}$ & -19.5 & -32.4 & -19.5 & -31.5 & -26.6 & -15.6 & -14.5 & \textbf{+13.8} & -17.2 \\
    \bottomrule
    \end{tabular}
    \caption{Networks trained with each method combined with our method on the Fashion-MNIST dataset with $\varepsilon=0.3$. Improved results are highlighted in bold. T*: TRADES+Ours, M*: Madry+Ours, R*: ReLU Stable+Ours.}
\label{tab: combine_fashion}\end{table*}

\begin{table*}[!t]
    \centering\small
    \begin{tabular}{c|l|ccc|ccc|ccc}
        \toprule
        \multicolumn{2}{c|}{Model} & \multicolumn{3}{c|}{C1} & \multicolumn{3}{c|}{C2} & \multicolumn{3}{c}{C3} \\ \hline
        \multicolumn{2}{c|}{Method} & T* & M* & R* & T* & M* & R* & T* & M* & R* \\ \hline
        \multicolumn{2}{c|}{Test Acc.} & -9.2\% & -14.2\% & -8.1\% & \textbf{+5.4\%}& -11.0\% & -0.1\% & -0.2\% & -9.3\% & -13.4\% \\ \hline
        \multirow{5}{*}{$\varepsilon$=$\frac{2}{255}$} & UNSAT\% & -6.0 & -7.5 & -7.5 & -9.5 & -4.0 & -2.5 & -0.9 & \textbf{+35.5} & \textbf{+32.5} \\
         & Stable\% & \textbf{+17.2} & \textbf{+17.8} & \textbf{+13.5} & \textbf{+14.3} & \textbf{+22.2} & \textbf{+17.2} & \textbf{+9.5} & \textbf{+45.7} & \textbf{+47.3} \\
         & Time (s) & \textbf{-3.9} & \textbf{-6.3} & \textbf{-3.5} & +8.2 & \textbf{-6.7} & \textbf{-11.0} & +3.5 & \textbf{-85.4} & \textbf{-84.1} \\
         & Time$_\text{U+T}$ (s) & \textbf{-6.2} & \textbf{-11.5} & \textbf{-6.8} & +19.4 & \textbf{-9.3} & \textbf{-17.1} & +4.7 & \textbf{-120.0} & \textbf{-123.1} \\
         & $\text{PGD}_{100}$ & -7.9 & -10.3 & -6.5 & -1.5 & -7.8 & -6.3 & -0.5 & -4.8 & -18.6 \\
        \hline
        \multirow{5}{*}{$\varepsilon$=$\frac{4}{255}$} & UNSAT\% & \textbf{+7.5} & \textbf{+15.0} & \textbf{+5.7} & -10.5 & \textbf{+17.5} & \textbf{+2.5} & \textbf{+0.0} & \textbf{+2.5} & \textbf{+5.0} \\       
         & Stable\% & \textbf{+42.7} & \textbf{+45.7} & \textbf{+35.3} & \textbf{+36.6} & \textbf{+52.4} & \textbf{+46.5} & \textbf{+15.2} & \textbf{+58.1} & \textbf{+76.1} \\
         & Time (s) & \textbf{-23.3} & \textbf{-38.5} & \textbf{-20.6} & \textbf{-14.1} & \textbf{-44.9} & \textbf{-31.6} & +1.0 & \textbf{-18.0} & \textbf{-46.5} \\
         & Time$_\text{U+T}$ (s) & \textbf{-58.6} & \textbf{-87.7} & \textbf{-48.5} & +20.4 & \textbf{-82.9} & \textbf{-48.9} & +3.1 & \textbf{-6.4} & \textbf{-25.5} \\
         & $\text{PGD}_{100}$ & -6.6 & -5.9 & -5.1 & -9.2 & -3.9 & -12.0 & -0.5 & \textbf{+0.3} & -13.9 \\
        \hline
        \multirow{5}{*}{$\varepsilon$=$\frac{6}{255}$} & UNSAT\% & \textbf{+13.5} & \textbf{+21.0} & \textbf{+15.5} & 0.0 & \textbf{+3.5} & \textbf{+0.5} & -1.0 & \textbf{+0.5} & \textbf{+10.6} \\
         & Stable\% & \textbf{+62.2} & \textbf{+59.6} & \textbf{+53.0} & \textbf{+47.7} & \textbf{+50.1} & \textbf{+56.0} & \textbf{+16.5} & \textbf{+51.6} & \textbf{+77.3} \\
         & Time (s) & \textbf{-22.8} & \textbf{-36.2} & \textbf{-35.4} & \textbf{-38.4} & \textbf{-5.2} & \textbf{-27.3} & \textbf{-2.0} & \textbf{-9.1} & \textbf{-33.0} \\
         & Time$_\text{U+T}$ (s) & \textbf{-85.2} & \textbf{-121.5} & \textbf{-99.7} & \textbf{-7.8} & \textbf{-15.4} & \textbf{-9.6} & +3.7 & 0.0 & \textbf{-48.3} \\
         & $\text{PGD}_{100}$ & -5.4 & -1.5 & -4.2 & -14.7 & -0.4 & -15.9 & -0.6 & \textbf{+4.9} & -8.8 \\
        \hline
        \multirow{5}{*}{$\varepsilon$=$\frac{8}{255}$} & UNSAT\% & \textbf{+9.5} & \textbf{+10.1} & \textbf{+13.0} & -0.5 & \textbf{+0.5} & -0.5 & 0.0 & 0.0 & \textbf{+2.0} \\
         & Stable\% & \textbf{+67.9} & \textbf{+57.6} & \textbf{+56.7} & \textbf{+50.8} & \textbf{+39.1} & \textbf{+52.2} & \textbf{+16.1} & \textbf{+34.0} & \textbf{+69.2} \\
         & Time (s) & \textbf{-16.6} & \textbf{-7.2} & \textbf{-22.4} & \textbf{-33.3} & +0.9 & \textbf{-31.3} & \textbf{-7.5} & +3.5 & \textbf{-13.9} \\
         & Time$_\text{U+T}$ (s) & \textbf{-69.5} & \textbf{-53.5} & \textbf{-81.8} & +3.2 & \textbf{-7.0} & +3.4 & 0.0 & 0.0 & \textbf{-12.2} \\
         & $\text{PGD}_{100}$ & -3.6 & \textbf{+2.0} & -2.7 & -17.7 & \textbf{+2.2} & -17.8 & -0.1 & \textbf{+7.9} & -3.2 \\
        \bottomrule
    \end{tabular}
    \caption{Networks trained with each method combined with our method on the 
    CIFAR-10 dataset with $\varepsilon=8/255$. Improved results are highlighted in bold. T*: TRADES+Ours, M*: Madry+Ours, R*: ReLU Stable+Ours.}
    \label{tab: combine_cifar}
\end{table*}

\begin{table*}[!t]
    \centering\small
    \begin{tabular}{c|c|c|ccccccc}
        \toprule
        Model & Method & Test Acc & UNSAT\% & \#UNSAT & Time & Time$_\text{U+T}$ & $\text{PGD}_{50}$ & $\text{PGD}_{100}$ & Stable\% \\
        \hline
    \multirow{7}{*}{C1} & TRADES & 69.8 & 38.5 & 77 & 43.4 & 67.3 & 55.8 & 55.8 & 53.0 \\ 
        & Madry & 69.6 & 39.5 & 79 & 41.7 & 63.8 & 54.3 & 54.3 & 50.8 \\ 
        & RS & 69.3 & 44.5 & 89 & 30.9 & 44.4 & \textbf{56.7} & \textbf{56.7} & 60.9 \\ 
        & Ours & 69.0 & \textbf{46.0} & \textbf{92} & \textbf{27.3} & \textbf{37.8} & 56.4 & 56.4 & \textbf{77.6} \\ \cline{2-10}
        & RS & 75.0 & 1.5 & 3 & \textbf{49.9} & 171.9 & 26.4 & 26.4 & 8.5 \\ 
        & Ours & 75.0 & \textbf{33.0} & \textbf{66} & 54.4 & \textbf{87.0} & \textbf{56.1} & \textbf{56.1} & \textbf{46.3} \\ \cline{2-10}
        & Natural & 77.6 & 0.5 & 1 & 49.6 & 178.4 & 10.0 & 10.0 & 5.3 \\ \cline{1-10}
    \multirow{7}{*}{C2} & TRADES & 74.7 & 35.0 & 70 & 55.0 & 85.3 & 60.4 & 60.4 & 46.9 \\ 
        & Madry & 74.4 & 28.5 & 57 & 65.5 & 103.1 & 58.7 & 58.8 & 39.4 \\ 
        & RS & 74.4 & 41.5 & 83 & 53.9 & 77.1 & \textbf{61.4} & \textbf{61.4} & 49.4 \\ 
        & Ours & 74.0 & \textbf{45.5} & \textbf{91} & \textbf{38.8} & \textbf{55.0} & 61.3 & 61.3 & \textbf{56.8} \\ \cline{2-10}
        & RS & 79.3 & 1.0 & 2 & \textbf{54.1} & 173.9 & 28.5 & 28.4 & 3.7 \\ 
        & Ours & 79.3 & \textbf{12.0} & \textbf{24} & 87.9 & \textbf{147.3} & \textbf{57.3} & \textbf{57.2} & \textbf{26.3} \\ \cline{2-10}
        & Natural & 80.6 & 0.5 & 1 & 53.3 & 181.0 & 10.0 & 10.0 & 2.8 \\ \cline{1-10}
    \multirow{7}{*}{C3} & TRADES & 77.6 & 5.5 & 11 & 112.8 & 165.9 & 64.0 & 64.0 & 27.7 \\ 
        & Madry & 77.8 & 1.0 & 2 & 113.9 & 177.2 & 60.1 & 60.1 & 17.0 \\ 
        & RS & 78.0 & 2.5 & 5 & 122.2 & 176.8 & 61.1 & 61.2 & 21.9 \\ 
        & Ours & 77.0 & \textbf{25.6} & \textbf{51} & \textbf{95.4} & \textbf{126.2} & \textbf{66.7} & \textbf{66.7} & \textbf{80.1} \\ \cline{2-10}
        & RS & 83.7 & 0.0 & 0 & \textbf{65.5} & 180.4 & 30.3 & 30.3 & 0.1 \\ 
        & Ours & 83.3 & \textbf{1.0} & \textbf{2} & 115.6 & \textbf{177.3} & \textbf{61.0} & \textbf{61.0} & \textbf{5.6} \\ \cline{2-10}
        & Natural & 84.2 & 0.0 & 0 & 68.0 & 184.1 & 10.0 & 10.0 & 0.0 \\ 
    \bottomrule
    \end{tabular}
    \caption{Networks trained on CIFAR-10 dataset with $\varepsilon=2/255$. The best results are highlighted in bold. 'Natural' denotes the accuracy with only CE loss used.}
    \label{tab: cifar_close_accuracy}
\end{table*}

\section{Detailed Experimental Results}
The following tables provide detailed results of the experiments conducted for RQ1, RQ2, and RQ3.
Typically, Table~\ref{tab: fashion_results}, \ref{tab: cifar_results} and Table~\ref{tab: combine_fashion}, \ref{tab: combine_cifar} show the results of networks trained for RQ1 and RQ2, on Fashion-MNIST and CIFAR-10 datasets, respectively. 
Table~\ref{tab: cifar_close_accuracy} shows the results of networks trained for RQ3 on CIFAR-10 datasets.

\subsection{Extra Results for RQ1}

Table~\ref{tab: fashion_results} shows the evaluation results of networks trained in Fashion-MNIST. 

In the M1 model, our method demonstrates moderate accuracy. Although our method is not the best in terms of UNSAT\%, it is within 1\% of the best performing RS method. Our method performs best in terms of UNSAT\% at $\varepsilon=0.3$. Our method consistently maintains the best performance in terms of stability, with a stable ratio of over 50\%. Our method is the fastest in terms of verification time, and its advantage becomes more pronounced as the perturbation radius increases. Our method performs moderately under the PGD attack.

In the M2 model, our method still performs best in terms of Stable\%. However, various methods have their advantages and disadvantages.

It should be noted that in the M3 model, our method performs best in terms of UNSAT\%, Stable\%, and the average time of UNSAT and timeout issues. Especially at $\varepsilon=0.3$, our method performs best in terms of UNSAT\%, reaching 17.4\%, while networks trained with other methods cannot verify any UNSAT properties. Our method performs moderately under the PGD attack.

Table~\ref{tab: cifar_results} shows the evaluation results of networks trained on CIFAR-10. 
Although our method does not perform well on test set accuracy compared to other approaches, it outperforms other metrics.

Our method demonstrates superior performance in terms of the average verification speed and the average speed of UNSAT and Timeout issues, requiring less verification time on almost all verification tasks. This is especially evident in the C1 and C2 models, where our method consistently outperforms other approaches in terms of verification time. Although the accuracy under 100 steps of the PGD attack is not the best-performing method, our method still achieves competitive PGD accuracy.

Our method outperforms others in terms of verification accuracy as adversarial perturbations increase, affirming its ability to maintain verification-friendly properties. Our approach consistently maintains a high ratio of stable neurons, even under the most challenging perturbation radii.
This is particularly evident in the C2 and C3 models, where our method achieves a stable neuron ratio of over 50\% at the highest perturbation radius. In contrast, other methods almost lose all stable neurons at the same perturbation radius.

\subsection{Extra Results for RQ2}

Table~\ref{tab: combine_fashion} and Table~\ref{tab: combine_cifar} show the results of networks trained with each method combined with our method on the Fashion-MNIST and CIFAR-10 datasets, respectively.

Our method significantly increases the ratio of stable neurons, which implies a smaller theoretical upper bound for solving. In most cases, the verification time is greatly reduced after combining our method. The UNSAT\% slightly decreases when combined with our method at the smallest $\varepsilon$, but as the robustness radius $\varepsilon$ increases, our method performs almost the best in terms of UNSAT\%. It is worth noting that, in the largest model, combining our method almost improves all metrics, and the decrease in test accuracy and PGD accuracy is not significant. Especially in the Fashion-MNIST dataset, after combining our method, the UNSAT\% of the RS method in M3, $\varepsilon=0.3$ increases from 0.0 to 30\%, which is a significant improvement considering the verification capability of the solver.

When combining our method with other training methods, the network's verification performance is significantly improved; however, the test accuracy and robustness may be reduced. It needs a trade-off between verification performance and robustness depending on the application scenario when combining our method with other training methods.
\begin{table*}[!t]
    \centering\small
    \begin{tabular}{c|c|c|cccccc}
        \toprule
        Model & $\varepsilon$ & $\gamma[i]$ &  Test Acc & UNSAT\% & Time & Time$_\text{U+T}$ & $\text{PGD}_{100}$ & Stable\% \\
        \hline
    \multirow{3}{*}{M1} & \multirow{3}{*}{0.1}& $1$ & 97.61 & 100 & 3.24 & 3.24 & 51.16 & 89.8\\
                        & & $2^{r[i]}$              & 98.61 & 100 & 3.22 & 3.22 & 52.18 & 85.9 \\
                        & & $r[i]\cdot m[i]$        & 98.88 & 98  & 3.05 & 3.12 & 53.66 & 75.3\\ \hline
    \multirow{3}{*}{C1} & \multirow{4}{*}{$\frac{2}{255}$} & $1$ & 68.10 & 36 & 23.56 & 42.40 & 95.46 & 66.55\\
                        &  & $2^{r[i]}$                           & 69.53 & 35 & 35.83 & 63.57 & 97.46 & 62.05 \\
                        &  & $r[i]\cdot m[i]$                     & 71.15 & 32 & 44.53 & 77.45 & 97.83 & 51.37\\
    \bottomrule
    \end{tabular}
    \caption{Networks trained with different $\gamma[i]$ on CIFAR-10 dataset and MNIST dataset.}
    \label{tab: gamma}
\end{table*}

\begin{table*}[!t]
    \centering\small
    \begin{tabular}{c|c|c|cccccc}
        \toprule
        Model & $\varepsilon$ & Method &Test Acc & UNSAT\% & Time & Time$_\text{U+T}$ & $\text{PGD}_{100}$ & Stable\% \\
        \hline
    \multirow{2}{*}{M1} & \multirow{2}{*}{0.1}& SABR & 89.25 & 22  & 26.89 & 20.74 & 25.91 & 84.16\\
                        &                     & NBC  & 97.61 & 100 & 3.22 & 3.22 & 97.47 & 89.78\\\hline
    \multirow{2}{*}{C1} & \multirow{2}{*}{$\frac{2}{255}$} &SABR  & 59.57 & 50 & 6.55 & 3.49 & 48.53 & 87.57\\
                        &                                  &NBC   & 62.48 & 54 & 6.69 & 5.02 & 51.55  & 79.60\\
    \bottomrule
    \end{tabular}
    \caption{Comparison of networks trained with NBC and SABR on CIFAR-10 dataset and MNIST dataset.}
    \label{tab: compare certified training}
\end{table*}

\subsection{Discussion for Hyperparameters}

\begin{table}[t]
    \centering
    \begin{tabular}{l|ccccc}
        \toprule
        \multicolumn{1}{c|}{$\beta$} &0.0 & 1.0 & 3.0 & 6.0 & 9.0 \\
        \hline
        Test Acc & 76.0 & 72.2 & 70.0 & 67.1 & 66.0 \\
        UNSAT\%  & 0 & 22.0 & 34.0 & 32.0& 32.0\\
        Stable\% & 4.8  & 42.1& 60.9 & 71.8 & 74.3 \\
        Time (s) & 44.4 & 45.4 & 39.9 & 23.7 & 19.4\\
        Time$_\text{U+T}$ (s) & 180 & 96.1 & 71.0 & 42.9 & 34.4 \\
        $\text{PGD}_{100}$ & 0.8 & 49.1 & 53.2 & 52.5 & 52.5 \\ 
        \bottomrule
    \end{tabular}
    \caption{Networks (C1 model) trained with NBC at $\varepsilon=2/255$ on CIFAR-10 dataset with different $\beta$.}
    \label{tab: beta}
\end{table}

Table \ref{tab: beta} shows the results of C1 networks trained with NBC at $\varepsilon=2/255$ on CIFAR-10 dataset with different $\beta$. We evaluated the models with 50 verification tasks.
Overall, as $\beta$ increases, the test accuracy decreases, UNSAT\% increases, stable neuron ratio increases, and verification time decreases, which is consistent with our expectations. The results show that a larger $\beta$ can improve the verification performance of the network, but it may reduce the accuracy of the network. Therefore, it is necessary to choose an appropriate $\beta$ according to the specific requirements of the application scenario.

Table \ref{tab: gamma} shows the results of networks trained with different $\gamma[i]$ on CIFAR-10 dataset and MNIST dataset. We evaluated the models with 50 verification tasks.
As discussed in the main paper, neurons in smaller layers, when behavior is consistent, may constrain subsequent layers through propagation. Moreover, layers near the input and output often have fewer neurons, applying constraints to these layers can directly affect the forward or backward propagation process and accelerate the convergence of target loss. In contrast, middle layers typically have more neurons and are responsible for extracting complex features; over-constraining these layers could adversely affect the model's expressive power. The results show that using a looser penalty on layers with more neurons may increase accuracy but decrease the proportion of stable neurons, which is consistent with our expectations. This indicates that it is necessary to choose an appropriate penalty term based on the scenario.

\subsection{Compare with Certified Training}

Table \ref{tab: compare certified training} shows the comparison of networks trained with NBC and SABR. We evaluated the models with 50 verification tasks.
In the MINST dataset, the network trained with NBC achieves better performance in all metrics compared to the network trained with SABR. 
In the CIFAR-10 dataset, the network trained with NBC achieves better performance in terms of Test accuracy, UNSAT\%, and  accuracy under the PGD attack. 
These results show that under the same network architecture, our method—despite its lower computational cost and ease of implementation—achieves performance comparable to, or even surpasses, that of certified training methods which demand more computational resources.

\end{document}